\documentclass[letterpaper]{article} 
\usepackage{aaai24}  
\usepackage{times}  
\usepackage{helvet}  
\usepackage{courier}  
\usepackage[hyphens]{url}  
\usepackage{graphicx} 
\usepackage{natbib}  
\usepackage{caption} 
\usepackage[ruled,vlined,linesnumbered]{algorithm2e}

\usepackage{color}

\usepackage{booktabs} 
\usepackage{amsmath,amsthm,amssymb}
\usepackage[ruled,vlined,linesnumbered]{algorithm2e}

\newtheorem{definition}{Definition}
\newtheorem{theorem}{Theorem}
\newtheorem{lemma}{Lemma}
\newcommand{\squishlist}{
 \begin{list}{$\bullet$}
  { \setlength{\itemsep}{0pt}
     \setlength{\parsep}{1.5pt}
     \setlength{\topsep}{1.5pt} 
     \setlength{\partopsep}{0pt}
     \setlength{\leftmargin}{1.5em}
     \setlength{\labelwidth}{1em}
     \setlength{\labelsep}{0.5em} } }

\newcommand{\squishend}{
  \end{list}  }
\newcommand\newsubcap[1]{\phantomcaption%
       \caption*{\figurename~\thefigure(\thesubfigure): #1}}

\let\oldnl\nl
\newcommand{\nonl}{\renewcommand{\nl}{\let\nl\oldnl}}

\usepackage{subcaption}

\title{Handling Long and Richly Constrained Tasks through Constrained Hierarchical Reinforcement Learning}
\author {
    Yuxiao Lu\textsuperscript{\rm 1},
    Arunesh Sinha\textsuperscript{\rm 2},
    Pradeep Varakantham\textsuperscript{\rm 1}
}
\affiliations {
    \textsuperscript{\rm 1}Singapore Management University\\
    \textsuperscript{\rm 2}Rutgers University\\
    yxlu.2021@phdcs.smu.edu.sg, arunesh.sinha@rutgers.edu, pradeepv@smu.edu.sg
}

\usepackage{bibentry}

\begin{document}

\maketitle

\begin{abstract}
Safety in goal directed Reinforcement Learning (RL) settings has typically been handled through constraints over trajectories and have demonstrated good performance in primarily short horizon tasks. In this paper, we are specifically interested in the problem of solving temporally extended decision making problems such as robots cleaning different areas in a house while avoiding slippery and unsafe areas (e.g., stairs) and retaining enough charge to move to a charging dock; in the presence of complex safety constraints.  Our key contribution is a  (safety) Constrained Search with Hierarchical Reinforcement Learning (CoSHRL) mechanism that combines an upper level constrained search agent (which computes a reward maximizing policy from a given start to a far away goal state while satisfying cost constraints) with a low-level goal conditioned RL agent (which estimates cost and reward values to move between nearby states). A major advantage of CoSHRL is that it can handle constraints on the cost value distribution (e.g., on Conditional Value at Risk, CVaR) and can adjust to flexible constraint thresholds without retraining. We perform extensive experiments with different types of safety constraints to demonstrate the utility of our approach over leading approaches in constrained and hierarchical RL.
\end{abstract}

\section{Introduction}
Reinforcement Learning (RL) is a framework to solve decision learning problems in environments that have an underlying (Partially Observable) Markov Decision Problem, (PO-)MDP. Deep Reinforcement Learning~\cite{franccois2018introduction,hernandez2019survey} approaches have been shown to solve large and complex decision making problems. For RL agents to be relevant in the day-to-day activities of humans, they need to handle a wide variety of temporally extended tasks while being safe. A few examples of such multi-level tasks are: (a) planning and searching for valuable targets by robots in challenging terrains (e.g., disaster areas) while navigating safely and preserving battery to reach a safe spot; (b) for autonomous electric vehicles to travel long distances in minimum time, they need to optimize the position of recharge locations along the way to ensure the vehicle is not left stranded; (c) cleaning robots to clean a house while avoiding slippery and unsafe areas (e.g., stairs) and retaining enough charge to move to a charging dock. The following key challenge needs to be addressed in the above mentioned problems of interest: 
\squishlist
    \item Computing an execution policy that satisfies safety constraints (in expectation or in a confidence bounded way) for temporally extended decision making problems in the presence of uncertainty. 
\squishend
Existing research in temporally extended decision making problem has focused on hierarchical RL methods~\cite{nachum2018data,zhang2020generating,kim2021landmark,levy2017learning}.  These approaches successfully solve long horizon tasks mainly in the widely applicable setting of goal conditioned RL~\cite{ijcai2022p770}, but they are unable to deal with safety constraints.  
On the other hand, most existing research in handling \emph{trajectory based safety constraints} has focused on constrained RL approaches~\cite{simao2021alwayssafe,gattami2021reinforcement}, where constraints are enforced on expected cost. A recent method that has considered percentile/confidence based constraints is WCSAC~\cite{yang2021wcsac}. Unfortunately, these constrained RL approaches are typically only able to solve short horizon problems where the goal is not too far away. 
We address the need to bring together these two threads of research on hierarchical RL and constrained RL, which have mostly progressed independently of each other ~\cite{roza2023safe}. 
 To that end, we propose a new Constrained Search with Hierarchical Reinforcement Learning (CoSHRL) approach, where there is a hierarchy of decision models:
(a) The lower level employs goal conditioned distributional RL to learn reward and cost distributions to move between two local states that are near to each other.  
(b) The upper level is a constrained search mechanism that builds on Informed RRT*~\cite{gammell2014informed} to identify the best waypoints to get from a given start state to a ``far'' away goal state. This is achieved while ensuring overall expected or percentile cost constraints (representative of robust safety measures) are enforced. 

\smallskip

\noindent \textbf{Contributions:}
Our key contributions are: (1) we provide a  \emph{scalable} \emph{constrained} search approach suited for \emph{long horizon tasks} within a hierarchical RL set-up, (2) we are able to handle rich \emph{percentile} constraints on cost  distribution, (3) the design of enforcing the constraints at the upper-level search allows \emph{fast recomputation} of policies in case the constraint threshold or start/goal states change, and 
(4) \emph{mathematical guarantee} for our constrained search method.  
Finally, we provide an extensive empirical comparison of CoSHRL to leading approaches in hierarchical and constrained RL. 

\smallskip

\noindent \textbf{Related Work:} \textit{Constrained RL} uses the Constrained MDP (CMDP) to maximize a reward function subject to \emph{expected} cost constraints~\cite{satija2020constrained,pankayaraj2023constrained,achiam2017constrained,gattami2021reinforcement,tessler2018reward,liang2018accelerated,chow2018lyapunov,simao2021alwayssafe, stooke2020responsive, liu2022constrained,yu2022towards,zhang2020first}. 
WCSAC~\cite{yang2021wcsac} extends Soft Actor-Critic and considers a certain level of CVaR of the cost distribution as a safety measure; \cite{chow2017risk} use Lagrangian approach for the same. \cite{sootla2022saute} prevent only worst case cost (no CVaR or expected) violation by tracking the cost budget in the state, which further does not allow for multiple constraints.
As far as we know and from benchmarking work~\cite{ray2019benchmarking}, there is no constrained RL designed for long-horizon tasks, and even for short-horizon all current approaches need retraining if the constraint threshold changes.

\textit{Hierarchical Reinforcement Learning} (HRL) addresses the problem of
sequential decision making 
at multiple levels of abstraction~\cite{kulkarni2016hierarchical,dietterich2000hierarchical}. The problem could be formulated with the framework of MDP and semi-MDP (SMDP)~\cite{sutton1999between}.
Utilizing off-policy RL algorithms, a number of recent methods such as HIRO~\cite{nachum2018data}, HRAC~\cite{zhang2020generating}, and HIGL~\cite{kim2021landmark} propose a hierarchy where both lower and upper level are RL learners and the higher level specifies sub-goals~\cite{kaelbling1993learning} for the lower level. 
However, it is hard to add safety constraints to such HRL with RL at both levels because to enforce constraints the higher level policy must generate constraint thresholds for the lower-level agent while ensuring the budget used by multiple invocations of the lower-level agent does not exceed the total cost budget. Also, the lower-level policy should be able to maximize reward for \emph{any} given cost threshold in the different invocations by the upper level. However, both these tasks are not realizable with the existing results in constrained RL. Options or skills learning coupled with a higher level policy of choosing options is another approach~\cite{eysenbach2018diversity,kim2019variational} in HRL. CoSHRL can be viewed as learning primitive skills of reaching local goals, and the simplicity of this task as well as of the search makes our approach scalable and flexible.

Closer to our method, SORB~\cite{eysenbach2019search} employs a graph-based path-planning (Dijkstra's algorithm) at the higher level and distributional RL at low level, where the continuous state is discretized to yield a massive graph. SORB achieves better success rate in complex maze environments compared to other HRL techniques but cannot enforce constraint and has high computational cost due to a large graph. We present a thorough comparison of our ConstrainedRRT* to SORB's planner in Section~\ref{sec:highlevel}. PALMER~\cite{beker2022palmer} employs RRT* for the high level, but instead of distributional RL at the low-level it uses an offline RL like approach, requiring a large pre-collected dataset fully covering the environment; importantly, PALMER also cannot enforce constraints. 

Logic based compositional RL~\cite{jothimurugan2021compositional,neary2022verifiable} shares similarities with our approach in terms of combining a high-level planner with a low-level RL agent. However, works in compositional RL have a binary logical specification of success, whereas we are in a quantitative setting of constrained MDP with rewards and cost constraints (and novel CVaR constraints). Also, our utilization of the
RRT* planner is quite different from the reachability planner used
in these works.

\subsection{Problem Formulation}
We have an agent interacting with an environment in a Markov Decision Process (MDP) setting. The agent observes its current state $s \in S$, where $S \subset \mathbb{R}^{d}$ is a continuous state space. The initial state $s_O$ for each episode is sampled according to a specified distribution and the agent seeks to reach goal state $s_G$. The agent's action space can be continuous ($a \subset \mathbb{R}^{n}$) or discrete. The episode terminates when the agent reaches the goal, or after $T$ steps, whichever occurs first. The agent earns immediate reward $r^t(s^t, a^t)$ and separately also incurs immediate cost $c^t(s^t, a^t)$ when acting in time step $t$. 
$V^\pi(s_O,s_G)$ and $V^\pi_c(s_O,s_G)$ are the cumulative undiscounted expected reward and cost respectively for reaching goal state $s_G$ from origin state $s_O$ following policy $\pi$.  The typical optimization in constrained RL~\cite{achiam2017constrained} is:
\begin{align}
\max_{\pi}  V^{\pi}(s_O,s_G) 
\quad s.t. \quad  V_c^\pi(s_O,s_G) \leq K 
\label{consMDP}
\end{align}
where the value functions are given as
$ \textstyle V^\pi(s_O,s_G) = \mathbb{E} \big[\sum_{t=0}^T r^t(s^t,a^t) | s^T = s_G, s^0 = s_O \big]$ and $ 
 \textstyle V^\pi_c(s_o,s_G) = \mathbb{E}\big[\sum_{t=0}^T c^t(s^t,a^t) | s^T = s_G, s^0 = s_O \big]
$ with the expectation taken over policy and environment. 

\begin{figure*}[t]
\centering
\includegraphics[width=0.9\linewidth]{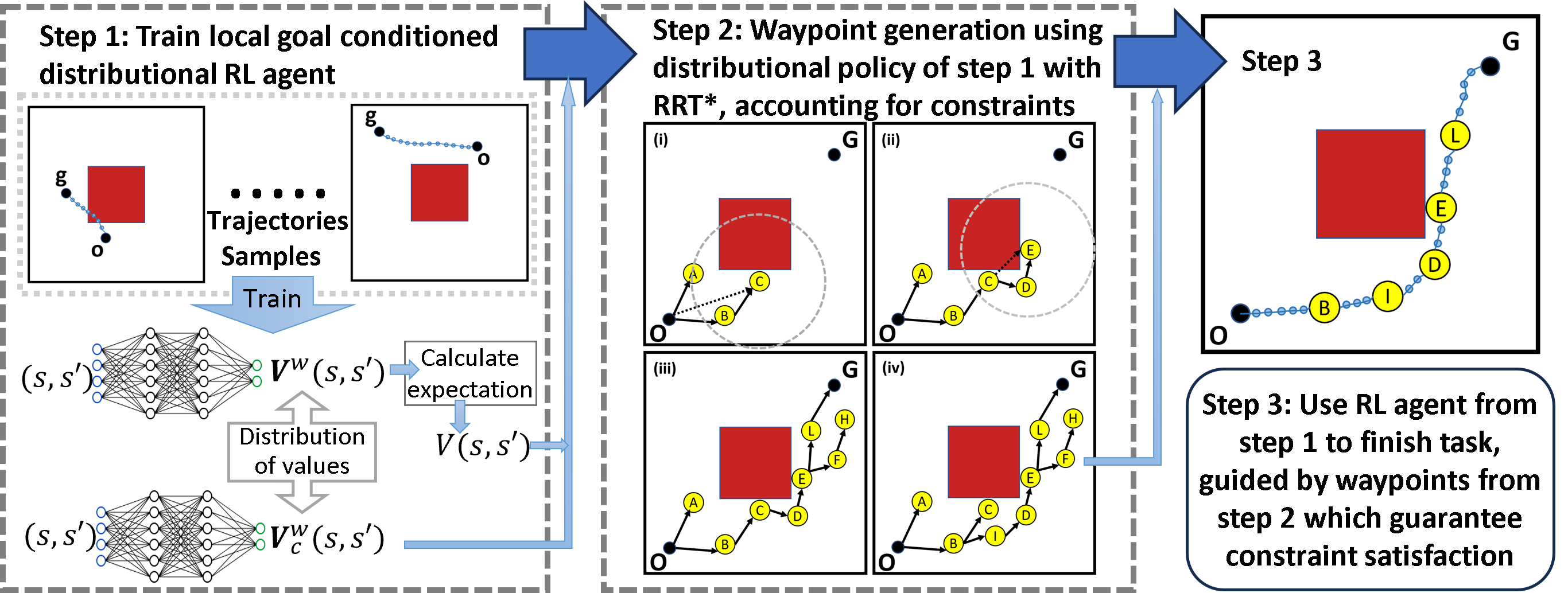}
\caption{Overview of CoSHRL.
\textbf{Step 1}: Train a local goal-conditioned RL agent using multiple randomly selected (o, g) (o is start, g is goal) pairs in a constrained environment (top part). The red square indicates a high-cost region. The learning is local and hence the goal will be unreachable if it's not ``near'' to the start. In this step, the local value function $V$ and the cost function $V_c$ are learned. \textbf{Step 2}: Generate waypoints guided by $V$ and $V_c$ using the proposed ConstrainedRRT* algorithm 
(i) The search samples state C, and O is not within the dashed circle of ``near'' states. Although both A and B are within the circle, the path from O to C via B is better as $V$(O,B) $+ V$(B,C) $< V$(O,A) $+ V$(A,C) using low-level agent's $V$ function. So, edge (B, C) is added to the tree. 
(ii) For new sample E, E is ``near'' from C and D, but the edge (C, E) is not valid because of cost constraint $CVaR_{\alpha}(\mathbf{V}_c$(O, B) $+ \mathbf{V}_c$(B, C) $+ \mathbf{V}_c$(C, E)$) > K$.
(iii) A path (O, B, C, D, E, L, G) within the cost constraint is found.
(iv) As the number of sampled states increases, a better path (O, B, I, D, E, L, G) is found. 
\textbf{Step 3}: Leveraging the waypoints from step 2, the pre-trained goal-conditioned RL agent completes the task.
}
\label{fig:overall}
\vskip -0.1in
\end{figure*}

However, in the above, the constraint on the expected cost value is not always suitable to represent constraints on safety. E.g., to ensure that an autonomous electric vehicle is not stranded on a highway, we need a robust constraint that ensures the chance of that happening is low, which cannot be enforced by expected cost constraint.  Therefore, we consider a cost constraint where we require that the CVaR~\cite{rockafellar2000optimization} of the cost distribution (given by the bold font random variable $\textbf{V}_c^\pi(s_O, s_G)$ is less than a threshold. We skip writing $s_O, s_G$ when implied. 
Intuitively, Value at Risk, $VaR_\alpha$ represents the minimum value for which the chance of violating the constraint (i.e., $\mathbf{V}_{c}^\pi > k$) is less than $\alpha$ specified as 
\begin{align*}
    VaR_\alpha(\textbf{V}_c^\pi)\! =\! \inf \{k ~|~ Pr(\mathbf{V}_{c}^\pi > k) \leq \alpha \}
    \end{align*}
Conditional VaR, $CVaR_\alpha$ intuitively refers to the expectation of values that are more than the $VaR_\alpha$, i.e., $CVaR_{\alpha}(\mathbf{V}_c^\pi)\! =\! \mathbb{E}[\mathbf{V}_c^\pi ~|~ \mathbf{V}_c^\pi \geq VaR_\alpha(\mathbf{V}_c^\pi) ]$. 
With this robust variant of the cost constraint (also known as percentile constraint), the problem that we solve for any given $\alpha$ is  
\begin{align}
    \textstyle \max_{\pi}  V^{\pi}(s_O,s_G) 
\quad s.t. \quad CVaR_{\alpha}(\mathbf{V}_c^\pi) \leq K \label{percentile}
\end{align}
Note that $\alpha\!=\!1$ is risk neutral, i.e., $CVaR_{1}(\mathbf{V}_c^\pi)\! =\! \mathbb{E}[\mathbf{V}_c^\pi ]\! =\! V_c^\pi$ , and $\alpha$ close to 0 is completely risk averse. 

\section{Approach}
\label{sec:approach}
Our approach, referred to as CoSHRL, solves the problem in Equation~\ref{percentile}. 
As shown in Figure~\ref{fig:overall}, CoSHRL employs a lower-level distributional RL agent and an upper-level search agent. First, the \emph{goal-conditioned}~\cite{kaelbling1993learning} off-policy distributional RL agent learns local distribution of reward and cost between states that are ``near'' to each other. Then, 
the upper-level agent is constructed using a constrained search algorithm by utilizing the reward and cost distributions. Finally, through its interactions with the environment, the lower-level agent reaches the far away goal guided by the waypoints produced by the constrained search. 

\smallskip

\noindent \textbf{Lower Level Agent}: Distributional RL~\cite{bdr2023} is a popular technique that enables learning distribution of value functions instead of just expected values. Distributional RL learns a policy $\hat{\pi}$ and maintains a network representing the distribution of $\mathbf{Q}$; we show how to derive $\mathbf{V}, \mathbf{V}_c$ from the learned policy $\hat{\pi}$.

\textit{Why distributional RL}? For rewards, we need to estimate just the expected $V^\pi(s,s')$, but it is known from the literature that learning the distribution of $\textbf{V}^\pi$ and then calculating expected value leads to better estimates~\cite{eysenbach2019search,beker2022palmer}. 
For completeness, we provide experimental evidence of this phenomenon in Appendix.
For enforcing percentile based cost constraint, we need to estimate the distribution of cost $\mathbf{V}_{c}$ for the $\hat{\pi}$ learned by lower-level agent. This is only possible with the use of distributional RL.

\textit{Representation}: In distributional RL for discrete actions, the distribution of $\mathbf{Q}$ is assumed to be over $N$ discrete values. The distribution of a goal conditioned $\mathbf{Q}$ is represented by $Q^\theta$ (neural network parameterized by $\theta$), which takes as input $s,s',a$ ($s'$ is local goal) and outputs a vector $[p_1, \ldots, p_{N}]$ where $p_i$ is the probability of expected reward value taking the $i^{th}$ discrete value. 
For completeness, the standard training of distributional RL is described in the Appendix, yielding a trained policy $\hat{\pi}$. 
For training, we choose nearby start and end states at random throughout the state space, relying on the generalizability of neural networks to obtain good estimates for nearby start and goal in the whole state space. 

Next, for discrete actions, we represent the distribution of value $\mathbf{V}^{\hat{\pi}}$ as a neural network $V^w$, which again outputs a probability vector. For simplicity, we do not include the learned policy $\hat{\pi}$ (which will not change) in the notation for $V^w$. The fixed learned $\hat{\pi}$ allows us to estimate $V^{w}$ directly by minimizing the KL divergence between a target $V^t(s,s') = Q^\theta(s,s',a), a\! \sim\! \hat{\pi}(\cdot|s,s')$ and the current $V^w$, i.e.,
    $\min_{w} D_{KL}(V^{t} || V^{w})$.
We optimize the above by storing experiences sampled according to $\hat{\pi}$ in a replay buffer and sampling mini-batches to minimize the loss above, analogous to supervised learning. Once the vector of probabilities $V^w$ is obtained, we can obtain the expected $V$ by calculating the expectation. 

For continuous actions, we can directly learn the distribution of $\mathbf{V}$, represented by a network $V^w$ using the same vector of probability representation of the distribution of value as used above for $Q^{\theta}$.

For problems in path search with no movement uncertainty, reward $r$ is set to $-1$ for each step such that the learned expected negated reward value function $-V(s,s')$ reflects the estimated length of the shortest path (avoiding impenetrable obstacles) from $s$ to $s'$ as done in~\cite{kaelbling1993learning,eysenbach2019search}. In particular, we assume that $-V$ is learned accurately and prove the following result:
\begin{lemma} \label{lemma:metric}
    Given $S \subset \mathbb{R}^d$, assuming $-V$ gives the obstacle avoiding shortest path length, $-V$ is a distance metric.
\end{lemma}
Next, for costs, we note that we performed the reward estimation without considering costs since in our approach the lower-level agent does not enforce constraints. However, the lower level agent does estimate the local costs as distributional $\mathbf{Q}$ values as a $Q^\theta_c$ network in the discrete action case or distributional $\mathbf{V}$ values as a $V^w_c$ network in the continuous action case. Then, in the discrete action case, similar to above learning of $V^w$, the fixed learned policy $\hat{\pi}$ allows us to estimate the vector of probability $V^w_c$ function directly by minimizing the KL divergence between a target $V_c^t = Q^\theta_c(s,s',a), a\! \sim\! \hat{\pi}(\cdot|s,s')$ 
and the current $V_c^w$: 
$
    \min_{w} D_{KL}(V_c^{t} || V_c^{w})
$. In the continuous action case,  the network $V^w_c$ is already learned directly
(details in Appendix).

\smallskip

\noindent\textbf{Upper Level Agent}:
\label{sec:highlevel}
Once the lower-level RL training is complete, we obtain a \emph{local} goal-conditioned value function for any origin and local goal state that are near to each other. In this section, we use the learned expected value $V$ and cost random variable $\mathbf{V}_c$ (\emph{removing superscripts for notation ease}).
First, we formulate the upper-level optimal constrained search problem. The RRT* search works in a continuous space $S \!\subset\! \mathbb{R}^d$. 
A path is a continuous function $\sigma: [0,1] \!\rightarrow\! \mathbb{R}^d$ with the start point as $\sigma(0)$ and end as $\sigma(1)$. In practice, a path is represented by a discrete number of states $\{\sigma(x_i)\}_{i\in[n]}$ for $0\!=\!x_0\! <\! x_1 \!<\! ... \!<\! x_{n-1} \!<\! x_{n}\!=\!1$ and some positive integer $n$ ($n$ can be different for different paths).
 A collision-free path is one that has no overlap with fixed obstacles. The set of all paths is $\Sigma$, and the set of obstacle free paths is $\Sigma_{free}$. A length of path is defined as $\sup_{n: 0=t_0 \!<\! \ldots t_n\!=\!1 }\sum_{i=1}^n d(x_{t_{i-1}}, x_{t_{i}})$ for given underlying distance $d$. The RRT* search (or the Informed version) finds the shortest path from the given start and end point.

 Given the discrete representation, for our CoSHRL the path traversed between $\sigma(x_i)$ and $\sigma(x_{i+1})$ is determined by the lower-level agent's policy. 
Every path $\sigma \in \Sigma$ provides a reward $R_{\sigma}$ and incurs a cost $C_{\sigma}$.  We define the reward for segment $(\sigma(x_i),\sigma(x_{i+1}))$ of a path as $V(\sigma(x_i),\sigma(x_{i+1}))$, where $V$ is the local goal-conditioned value function learned by the lower-level agent. Similarly, the cost incurred for segment $(\sigma(x_i),\sigma(x_{i+1}))$ is $\mathbf{V}_c(\sigma(x_i),\sigma(x_{i+1}))$. Thus, 
\begin{align}
    R_{\sigma} = \sum_{i=0}^{n-1} V(\sigma(x_i),\sigma(x_{i+1}))  \\  
    \mathbf{C}_{\sigma} = \sum_{i=0}^{n-1} \mathbf{V}_c(\sigma(x_i),\sigma(x_{i+1}))
\end{align}
In CoSHRL, the constrained search problem is to find the optimal path, $\sigma^*$ ($\in\! \arg\!\max_{\sigma \in \Sigma} R_{\sigma}$) from $s_{O}$ to $s_G$ subject to a cost threshold, i.e., $CVaR_\alpha( \mathbf{C}_{\sigma^*}) \leq K$. As $-V$ is the shortest distance considering obstacles (see the text before Lemma~\ref{lemma:metric}), the above optimization essentially finds the shortest path measured in distance $-V$ from $s_O$ to $s_G$ avoiding all obstacles and within the cost constraint $K$.

\IncMargin{0.0em}
\begin{algorithm}[t]
\SetAlCapHSkip{.0em}
\caption{ConstrainedRRT* ($s_{o}, s_{G}, V, \mathbf{V}_c, K$)}
    \label{algo:pathplanning}
    \DontPrintSemicolon
    $\mathcal{V} \gets \{s_{o}\}$, $\mathcal{E} \gets \emptyset $ , $S_{soln} \gets \emptyset$,   $\mathcal{T} = (\mathcal{V},\mathcal{E})$ \;
    \For{\rm{iteration} = 1 ... N}{
        $\mathcal{T}$ = \rm{${\sf Extend\_node}(s_{o}, s_{G}, V, \mathbf{V}_c, K, \mathcal{T}, S_{soln})$}
    }
    \Return best solution in $S_{soln}$ \;
\smallskip
\SetKwFunction{proc}{Extend\_node}
\SetKwProg{myproc}{def}{}{}
\nonl \myproc{\proc{$s_{o}, s_{G}, V, \mathbf{V}_c, K, \mathcal{T}, S_{soln}$}}{
Sample $s_{new}$ within $\min(r_{RRT*}, \eta)$ from its nearest node in $\mathcal{T}$ as in Informed RRT* \;
            $S_{near} \gets$ Find all nodes in $\mathcal{T}$ within $\min(r_{RRT*}, \eta)$ from $s_{new}$\;
            Find $s_{min} \in \arg\!\min_s \{-R(s)\! -\! V(s, s_{new})~\vert~ s \!\in\! S_{near}, $\rm{Valid\_edge}$(\mathcal{T}, s, s_{new}, \mathbf{V}_c, K)\}$ \; 
            $\mathcal{V} \gets \mathcal{V} \cup \{s_{new}\}$, $\mathcal{E} \gets \mathcal{E} \cup \{(s_{min}, s_{new})\}$ \;

            $S_{cand} = \{ s\! \in\! S_{near} ~\vert~ -R(s_{new})\! - V(s_{new}, s) \!<\! -R(s),$ \rm{Valid\_edge}$(\mathcal{T}, s_{new}, s, \mathbf{V}_c, K) \}$ \;
            
            \For{$\forall s \in S_{cand}$}
            { 
                $s_{parent} \gets $Parent$(s)$, 
                $\mathcal{E} \gets \mathcal{E} \backslash \{(s_{parent}, s)\}$, 
                $\mathcal{V} \gets \mathcal{V} \cup \{s_{new}\}$, 
                $\mathcal{E} \gets \mathcal{E} \cup \{(s_{new},s)\}$    
            }
            If $s_{new}$ is near the goal $s_G$, then form $\sigma$ by tracking parents of $s_{new}$ and $S_{soln}\! \gets\! S_{soln} \!\cup\! \{\sigma\}$\;
}
\end{algorithm}

Our approach has immediate advantages over the state-of-the-art SORB~\cite{eysenbach2019search}, which also employs an upper level planner and lower level RL. SORB constructs a complete graph and then computes the shortest path using Dijkstra's algorithm.  However, SORB has fundamental limitations:
(1) The graph is built from the replay buffer of explored nodes. This can result in bad distribution of nodes in the state space (without considering start, goal, or obstacles).
(2) The coarse discretization can result in a non-optimal path between the start and goal state~\cite{karaman2011sampling}.   
(3) Construction of complete graph yields $O(N^2)$ complexity for Dijkstra's algorithm with $N$ nodes (compared to $N \log N$ for our search).

Thus, an online search method that samples and grows a tree from the given start to the goal state while avoiding extending into obstacles is more suited as the upper-level search. Hence we provide Constrained-RRT*, which builds on Informed-RRT*~\cite{gammell2014informed} to handle constraints. Informed-RRT* builds upon RRT*~\cite{karaman2011sampling}, which works by constructing a tree whose root is the start state and iteratively growing the tree by randomly sampling new points as nodes till the tree reaches the goal. In Informed RRT*, as an informed heuristic, the sampling is restricted to a \emph{specially constructed ellipsoid}. However, both Informed-RRT* and RRT* do not take constraints into account.

\begin{algorithm}[t]
\caption{Valid\_edge ($\mathcal{T}, s, s', \mathbf{V}_c, K$)}\label{algo:validcheck}
\DontPrintSemicolon
    $result \gets \mathbf{V}_c(s, s')$ \;
    \While{$s.parent$}{
        $result \gets result + \mathbf{V}_c(s.parent, s)$\;
        $s \gets s.parent$ \;
    }
    \If{$CVaR_{\alpha}(result) \leq K$}{
        \Return True
    }
    \Return False
\end{algorithm}

\noindent \textit{Algorithm Description}:
We propose Constrained RRT* (Algorithm~\ref{algo:pathplanning}), which builds on Informed RRT* to handle the cost constraint. The pseudocode is provided in Algorithm~\ref{algo:pathplanning}.  We search for the optimal path $\sigma^{*}$ by incrementally building a tree $\mathcal{T}$ in the state space $S$. The tree, $\mathcal{T}$ consists of a set of nodes, $\mathcal{V}$ ($\subset S$), and edges $\mathcal{E}$ ($ \subset S \times S$). 
In the sub-routine $\mathsf{Extend\_node}$, a candidate state $s_{new}$ is chosen (line 5) to be added to the tree $\mathcal{T}$ by a \emph{sampling process that is the same as in Informed RRT*} (see Appendix for details of sampling).
The hyper-parameter $\eta$ accounts for the fact that our distance estimates are precise only locally (see Appendix for hyperparameter settings). 

The rewiring radius, $r_{RRT*} = \gamma_{RRT*} (\log n/n)^{1/d}$, where $n$ is the current number of nodes sampled, is described in \cite{karaman2011sampling}. The node $s_{min}$ (line 7) that results in the shortest path (highest reward) to $s_{new}$ among the nearby nodes $S_{near}$ (line 6) is connected to $s_{new}$ in line 8, if the edge is valid.

Here, we take a detour to explain how we determine the validity of edges. An edge is valid if and only if adding it does not result in a (partial) path that violates the cost constraint. The key insight is that this validity can be determined by computing the convolution of the distributions associated with the (partial) path and the current $\mathbf{V}_c$. By providing the definition of \textcolor{black}{Valid\_edge ($\mathcal{T}, s, s', \mathbf{V}_c, K$)} in Algorithm~\ref{algo:validcheck} and doing the \textcolor{black}{Valid\_edge} checks in the $\mathsf{Extend\_node}$ subroutine, we ensure that any path output by the overall algorithm will satisfy the cost constraints.
In the pseudocode of \textcolor{black}{Valid\_edge}, $\mathbf{V}_c$ represents a random variable (and so does $result$). Then, the addition in line 3 of \textcolor{black}{Valid\_edge} is a convolution operation (recall that the distribution of a sum $\mathbf{X}+\mathbf{Y}$ of two random variables $\mathbf{X},\mathbf{Y}$ is found by a convolution~\cite{ross2014introduction}). 

Coming back to $\mathsf{Extend\_node}$, we explore further the possible edges to be added to the tree. In particular, in line 9 (1) the edge is created only if it is valid and (2) new edges are created from $s_{new}$ to vertices in $S_{near}$, if
the path through $s_{new}$ has lower distance (higher reward) than the path through the current parent; in this case, the
edge linking the vertex to its current parent is deleted, to maintain the tree structure. An example search run is shown in Figure~\ref{fig:overall}.

\smallskip

\textbf{Theoretical Results}:
The RRT* algorithm~\cite{karaman2011sampling} satisfies two properties: \emph{probabilistic completeness} and \emph{asymptotic optimality}. Intuitively,  probabilistic completeness says that as number of samples $n \!\rightarrow\! \infty$, RRT* finds a feasible path if it exists and asymptotic optimality says that as $n \!\rightarrow \! \infty$, RRT* finds the optimal path with the highest reward. Unsurprisingly, asymptotic optimality implies probabilistic completeness. Our key contribution is proving asymptotic optimality of ConstrainedRRT*, which requires complicated analysis because of constraints. 

\textit{Background}: We summarize many definitions from \citet{karaman2011sampling}. For detailed definition statements, we request the reader to peruse the referred paper.
\citet{karaman2011sampling} define addition and multiplication operations that make the set of paths $\Sigma$ a vector space. Further, they define a norm $||\sigma||_{BV}$ on this vector space (please refer to page 22 of~\cite{karaman2011sampling}). The distance induced by the BV norm allows for defining limits of a sequence of path, i.e., $\lim_{n \to \infty} \sigma_n$. A solution path $\sigma^*$ is called \emph{robustly optimal} if under the metric induced by the BV norm for any sequence of collision-free paths $\sigma_n$, if $\lim_{n \to \infty} \sigma_n = \sigma^*$ then $\lim_{n \to \infty} R_{\sigma_n} = R_{\sigma^*}$. A path is said to have \emph{strong $\delta$ clearance} if it is not within $\delta$ distance of any obstacle. A path $\sigma$ has weak $\delta$ clearance if there exists a sequence of paths with strong clearance converging to $\sigma$. For any path finding algorithm $ALG$, let $Y^{ALG}_n$ be the random variable corresponding to the reward
of the max-reward solution returned at the end of iteration $n$.
\begin{definition} [Asymptotic optimality~\cite{karaman2011sampling}]
\label{def:background}

An algorithm ALG is asymptotically optimal if, for
any path search problem that admits a robustly
optimal solution with finite reward $R^*$,
$
\mathbb{P}(\{\limsup_n Y^{ALG}_n = R^*\})\! =\! 1
$.
\end{definition}
\textit{Theoretical Results for Constraints}:
In this paper, due to the presence of constraints, we have to modify definitions. For instance, 
robustly optimal definition has to account for costs, i.e., the solution path $\sigma^*$ is called \emph{robustly optimal with constraints} if under the metric induced by the BV norm for any sequence of collision-free paths $\sigma_n$ if $\lim_{n \to \infty} \sigma_n = \sigma^*$ then $\lim_{n \to \infty} R_{\sigma_n} = R_{\sigma^*}$ and if $\lim_{n \to \infty} \sigma_n = \sigma^*$ then $\lim_{n \to \infty} C_{\sigma_n} = C_{\sigma^*}$. 
Next, the definition of weak $\delta$ clearance of optimal path $\sigma^*$ is extended to assume that there exists a sequence of strong $\delta$ clearance paths with total cost $\leq K + \epsilon$ when the path $\sigma^*$ has cost $\leq K$ for any small $\epsilon > 0$. Intuitively, this means that if the optimal path has cost at most $K$ then nearby strong $\delta$ clearance paths converging to the optimal path are also cost bounded closely by $K$ while allowing $\epsilon$ extra cost for possibly slightly longer paths.
We redefine Definition~\ref{def:background} with the cost constraint. Let $Z^
{ALG}_n$ be the random variable corresponding to the cost
of the max-reward solution included in the graph returned by $ALG$ at the end of iteration $n$ ($n$ samples). Then, we define:

\begin{figure*}
\centering
\begin{subfigure}[b]{0.32\textwidth}
\centering
\includegraphics[width=0.39\textwidth]{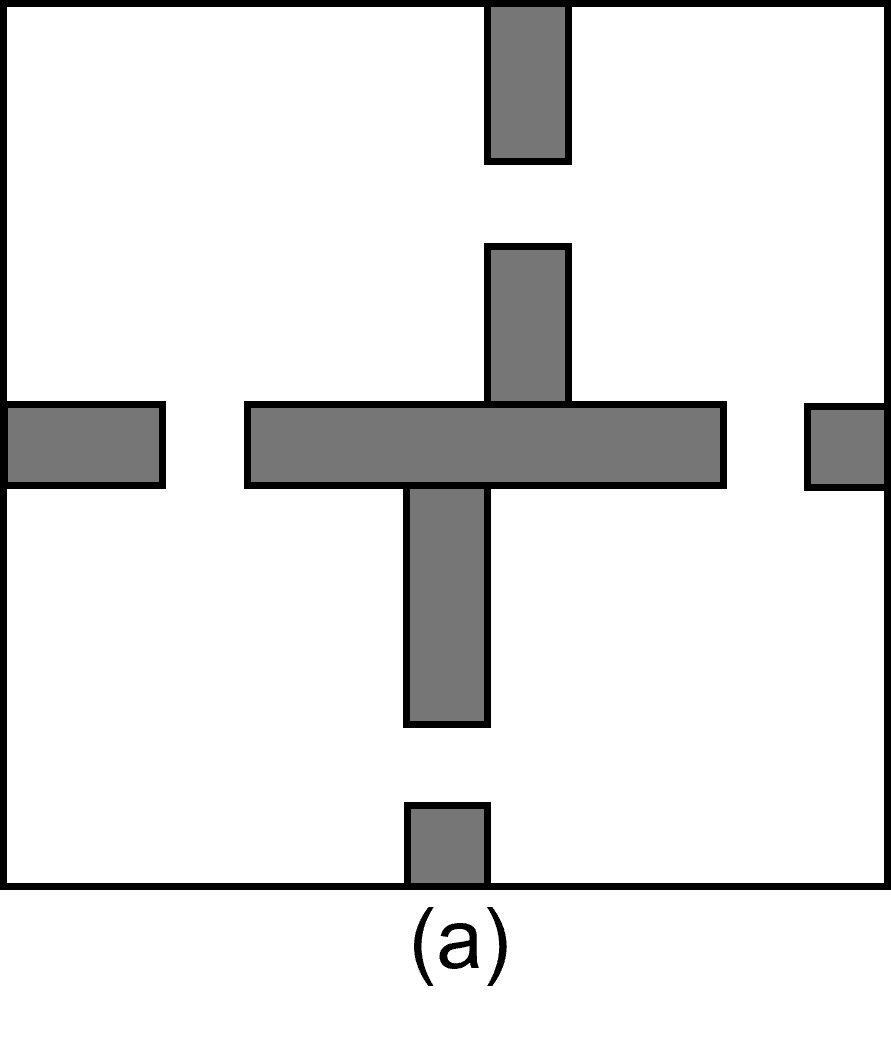}
\includegraphics[width=0.39\textwidth]{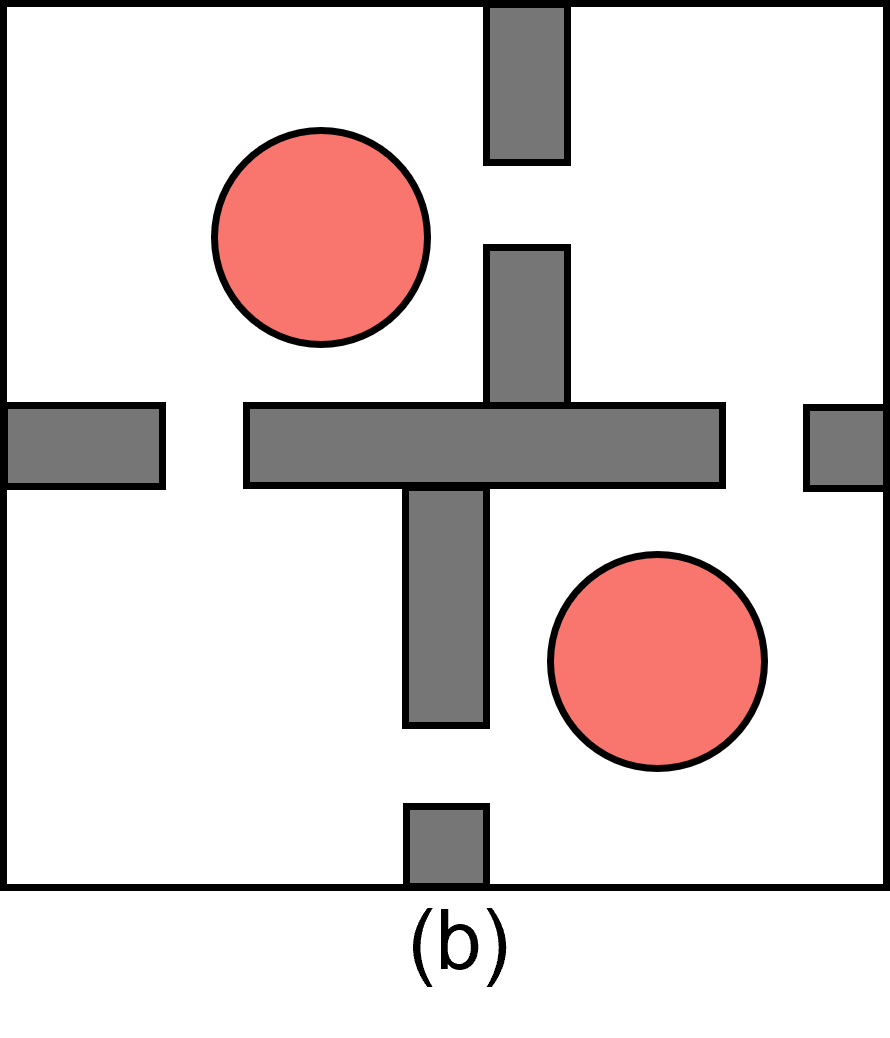}
\newsubcap{The complex point maze environment. Wall obstacles are in black. The environment on the right has hazardous red circles.}
\label{fig:fourroom}
\end{subfigure}
\hfill
\begin{subfigure}[b]{0.32\textwidth}
\centering
    \includegraphics[width=0.79\textwidth]{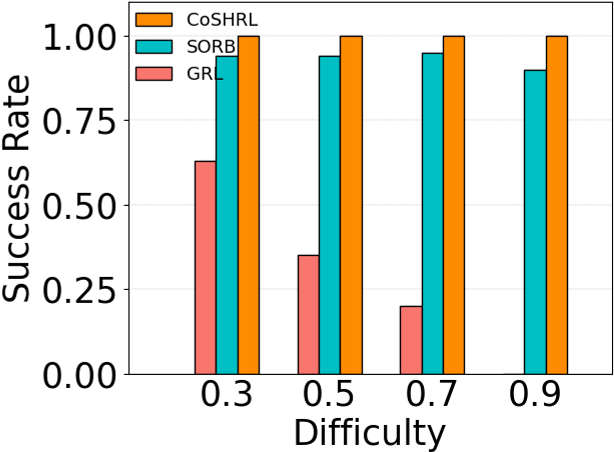}
    \newsubcap{Success rate vs Difficulty}
    \label{fig:success_rate}

\end{subfigure}
\hfill
\begin{subfigure}[b]{0.32\textwidth}
    \centering
    \includegraphics[width=0.79\textwidth]{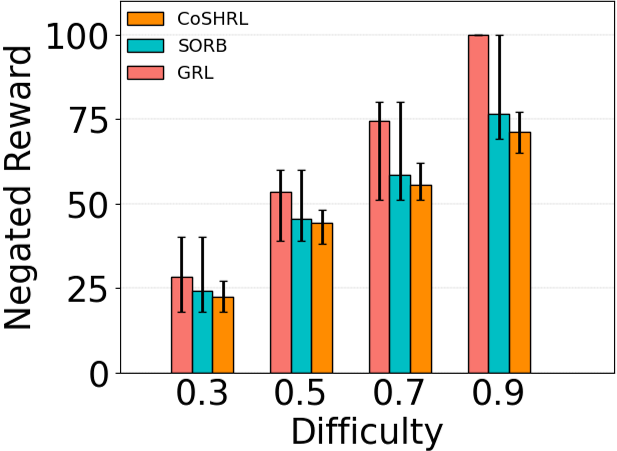}
    \newsubcap{Neg. reward vs Difficulty}
    \label{fig:reward_errorbar}
\end{subfigure}
\end{figure*}

\begin{definition}[Asymptotic optimality with constraints] An algorithm ALG is asymptotically optimal with constraints if, for
any path search problem that admits a robustly
optimal solution with finite cost constraints $K$ and with finite reward $R^*$,
$
\mathbb{P}(\{\limsup_n Y^{ALG}_n = R^*\}) = 1 \mbox{ and } Z^
{ALG}_n \leq K
$.
\end{definition}
We justify this definition as follows: since $ALG$ will stop in finite $n$, we require that the output of $ALG$ is always within the cost threshold $K$ for any $n$ at which the algorithm stops.
We prove that our change (Valid\_Edge check) preserves asymptotic optimality with constraints.
\begin{theorem} \label{thm:asymptotically_optimal}
Let $d$ be the dimension of the space $S$, $\mu(S_{free})$
denotes the Lebesgue measure (i.e., volume) of the obstacle-free space, and $\tau_d$ be the volume of the
unit Euclidean norm ball in the $d$-dimensional space.
    The Constrained RRT* in Algorithm~\ref{algo:pathplanning} preserves asymptotic optimality with constraints for $\gamma_{RRT*} \geq (2(1 + 1/d))^{1/d} \big(\frac{\mu(S_{free})}{\tau_d} \big)^{1/d}$.
\end{theorem}
The proof of RRT* involves constructing a random graph via a marked point process that is shown as equivalent to the RRT* algorithm. In the full proof in Appendix.
we incorporate cost constraints in the construction of the random graph and show its equivalence to ConstrainedRRT*. Then, the analysis is done for this constructed random graph.
The analysis involves (1) constructing a sequence of paths $\sigma_n$ with strong $\delta_n$ clearance converging to the optimal path $\sigma^*$ within cost constraint, (2) constructing a covering of the path $\sigma_n$ with a sequence of norm balls with radius $\delta_n/4$; we use a special value for $\delta_n$ to account for cost constraints. It is shown that with large enough $n$ and our special choice of $\delta_n$, the tree in ConstrainedRRT* will have a path satisfying cost constraints through these balls and will converge to $\sigma^*$.

\begin{figure*}
\centering
\begin{subfigure}[b]{0.32\textwidth}
    \centering
    \includegraphics[width=0.92\textwidth]{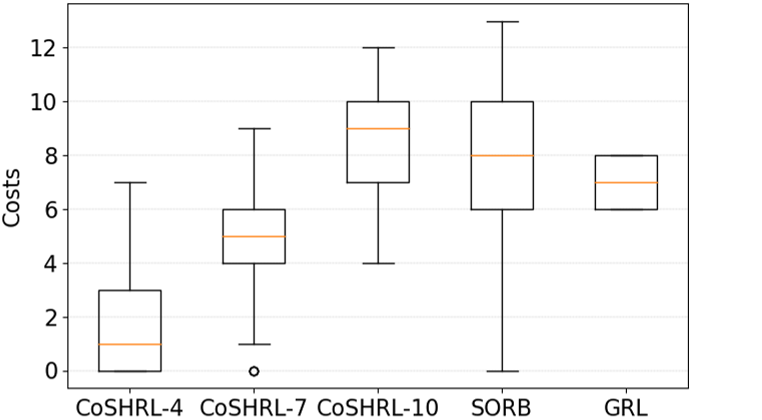}
\newsubcap{Boxplot of cost in evaluation for static risk after training in maze environment. 
}
\label{fig:fourroom_cost}
\end{subfigure}
\hfill
\begin{subfigure}[b]{0.34\textwidth}
    \centering
    \includegraphics[width=0.99\textwidth]{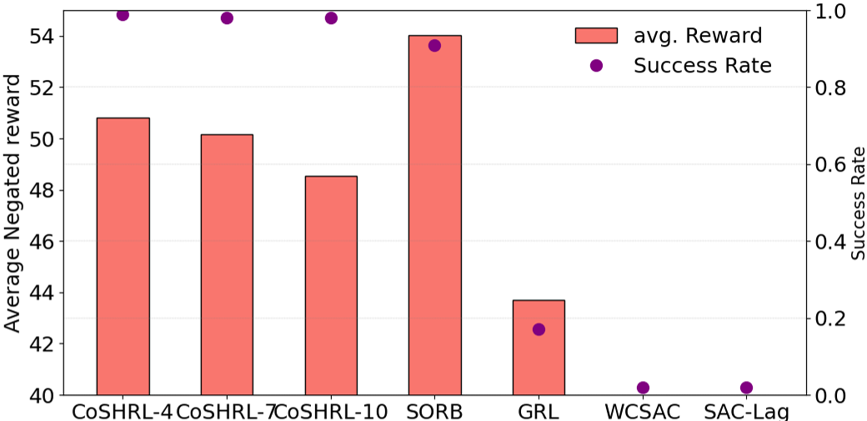}
\newsubcap{Success rate and avg. neg. reward of our method and baselines in maze environment. Only successful trials are counted for reward.}
\label{fig:fourroom_success_rate_reward}
\end{subfigure}
\hfill
\begin{subfigure}[b]{0.32\textwidth}
    \centering
    \includegraphics[width=0.92\textwidth]{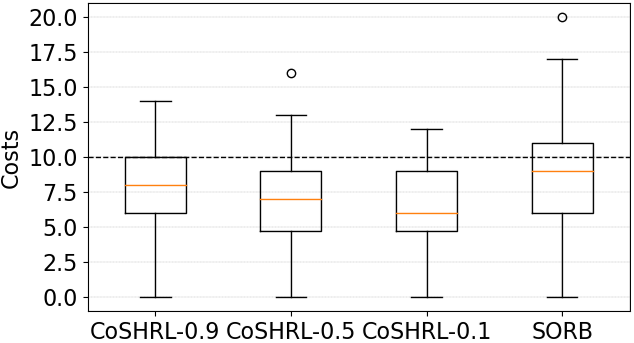}
        \newsubcap{Boxplot of cost evaluation for stochastic risk after training}
        \label{fig:fourroom_stochastic_cost}
\end{subfigure}
\end{figure*}

\section{Experiments}
We evaluate our method on two complex point maze environments and a novel image-based ViZDoom environment which have been used as a benchmark in RL navigation tasks~\citep{zhang2020generating,nachum2018data,beker2022palmer}. These maps include \emph{obstacles} (impenetrable) and \emph{hazards} (high cost but penetrable). We compare against SAC-Lagrangian (SAC-lag)~\cite{yang2021wcsac,stooke2020responsive}, WCSAC~\cite{yang2021wcsac}, SORB~\cite{eysenbach2019search}, and Goal-conditioned RL (GRL)~\cite{kaelbling1993learning}. SORB and GRL are not designed to enforce constraints, so they can get higher rewards but suffer from constraint violations.
\emph{Hyperparameter settings and additional results on other environments are in Appendix}.

\noindent \textbf{2D Navigation with Obstacles}: The first environment is point maze environment of Figure~\ref{fig:fourroom} (left), which has wall obstacles, but no hazards (thus, no cost constraints). The start point is randomly set in the environment while the goal is set $69\nu$ away from the start where $\nu$ is the difficulty level. As the immediate reward $r(s, a)\!=\!-1$, the agent needs to reach the goal using the shortest path that avoids the walls. 

We compare CoSHRL with goal-conditioned RL and SORB at different difficulty levels. For a fair comparison, both the number of nodes for SORB and the number of iterations for our method are set as 1000. For each experiment, we ran 100 trials with different seeds. We compare (a) the percentage of times the agent reaches the goal; and (b) the negated reward (i.e., the path length). 
In Figure~\ref{fig:success_rate}, we observe that the success rate of CoSHRL is 100\% and it outperforms SORB with a larger margin as the difficulty level increases. In Figure~\ref{fig:reward_errorbar}, 
we show all trials' negated reward (lower is better) for GRL, SORB, and CoSHRL.
The difficulty level $\nu$ decides the optimal distance between start and goal, e.g., when $\nu = 0.3$, the optimal distance is set at $69\times 0.3 \approx 21$; we observed that the baseline approaches frequently provided very circuitous paths much longer than the optimal path, e.g., SORB and GRL often provide circuitous paths with length exceeding 40 for $\nu=0.3$, so we cut them off at 40 for $\nu = 0.3$. We cut all trajectories off for baselines (thereby providing advantage to baselines) at $40, 60, 80, 100$ for difficulty levels $0.3, 0.5, 0.7, 0.9$ respectively. 

Yet, we observe that not only the average negated reward (path length) but also the upper bound and lower bound outperform SORB and GRL at different difficulty levels.

\noindent \textbf{2D Navigation with Obstacles and Hazards}:
In this part, we evaluate our method in the point maze environment of Figure~\ref{fig:fourroom} (right), where there are two hazards set in the top left room and bottom left room. The agent starts randomly in the bottom left room and the goal is randomly set in the top right room. The trajectory length will be longer if the agent tries to avoid the hazardous area.  We show results for static costs as well as for stochastic costs at different risk levels. It is worth noting that we don't need to retrain our lower-level RL policy for different cost thresholds $K$.

\begin{figure*}[t]
    \centering
    \begin{subfigure}[b]{0.55\textwidth}
    \centering
    \includegraphics[width=0.95\textwidth]{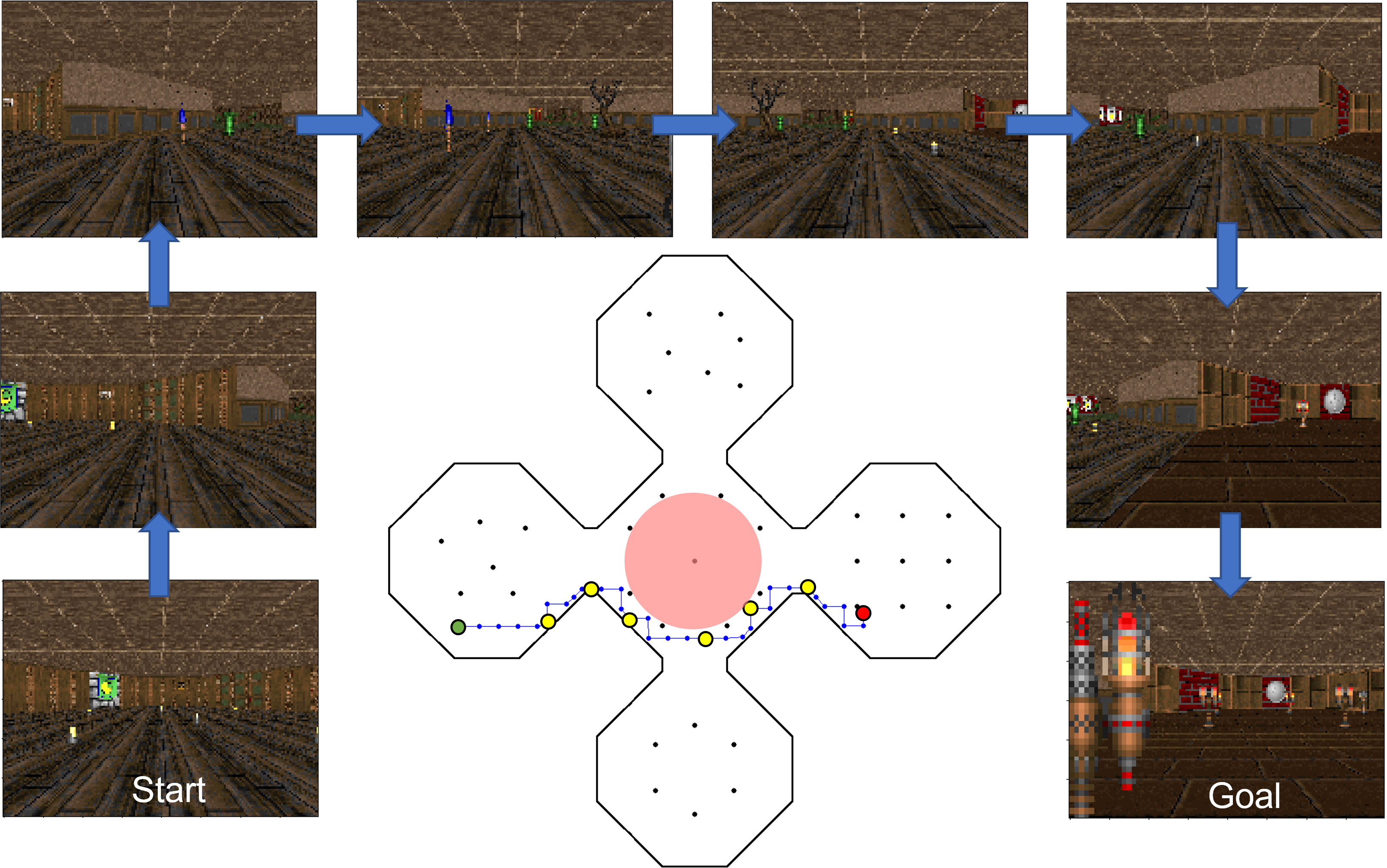}
        \newsubcap{An example safe trajectory in the safe ViZDoom environment. Fixed obstacles are shown in black points and hazardous area is shown in red circle. Given a start state (green point) and goal state (red point), our method could find a sequence of waypoints (yellow points) conditioning on flexible constraints threshold $K$ ($K=0$ in this figure). Using the low level RL between the waypoints our method could reach the goal constraints (shown in the blue line).}
        \label{fig:ViZDoom}
    \end{subfigure}
        \hfill
        \begin{subfigure}[b]{.4\textwidth}
        \centering
        \includegraphics[width=0.75\textwidth]{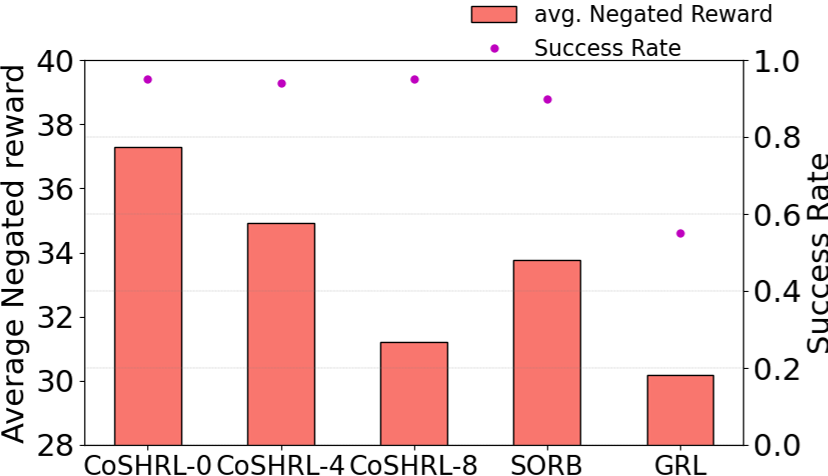}
        \newsubcap{Success rate and avg. negated reward of our method, SORB, and GRL in Safe-ViZDoom. Only successful trials are counted for reward}
        \label{fig:ViZDoom_success_rate_reward}
        \centering
        \includegraphics[width=0.75\textwidth]{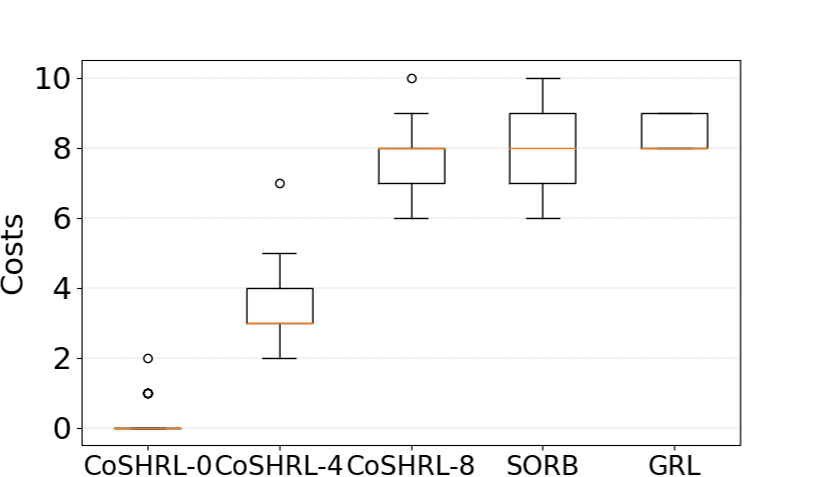}
        \newsubcap{Boxplot of cost in evaluation after training in Safe-ViZDoom.}
        \label{fig:ViZDoom_cost}
        \end{subfigure}
\end{figure*}

\noindent \emph{Static Cost}:
\label{sec:fourroom_cost}
In this environment, the agent incurs a cost $c\! =\! 1$ for each step in the hazard, otherwise $c = 0$. We evaluate our method with different cost limits $K$ shown with CoSHRL-4, CoSHRL-7, and CoSHRL-10 in Figure~\ref{fig:fourroom_success_rate_reward} and Figure~\ref{fig:fourroom_cost}.
In Figure~\ref{fig:fourroom_success_rate_reward}, the bars provide the path length (negated reward) to reach the goal (plotted on the primary Y-axis) and the purple dots indicate the success rate (plotted on the secondary Y-axis). For average negated reward (path length), we only consider the successful trials for all algorithms. We have the following key observations from Figure~\ref{fig:fourroom_success_rate_reward}: (1) Our method reaches the goal with a high success rate under different cost limits with nearly 100\% success. 
(2) Even though our method considers cost constraints, it is able to outperform SORB (which does not consider the cost constraint) not only in success rate but also in the length of the trajectory (average negated reward). 
(3) The success rate of GRL (goal-conditioned RL) is less than 20\% but for the average negated reward we only count the successful trials, hence the negated reward for goal-conditioned RL is better (lower) than our method. 
(4) WCSAC and SAC-Lagrangian, both non-hierarchical RL techniques that consider cost constraints, have $\approx\! 0$\% success rate in this long-horizon task and we don't consider them as baselines in further experiments.

The min., max. and mean cost for the different algorithms are shown in Figure~\ref{fig:fourroom_cost}. With increasing cost limit, the upper bound, lower bound, and median of the total cost increase for CoSHRL. This is expected as the path in the hazardous area increases and therefore potentially the error in the computation of $\mathbf{V}_c$ can increase. The proportion of trajectories that exceed the cost limit $K = 4, 7, 10$ are $4\%$, $6\%, 6\%$ respectively. 
Examples of paths produced by different approaches are shown in Appendix.

\noindent \emph{Stochastic Cost}:
In this environment, the agent incurs a cost $c$ uniformly sampled from $\{0, 1, 2\}$ for each step in the hazard, otherwise $c = 0$, i.e., the total cost of $n$ steps inside the hazard follows a multinomial distribution. In safety-critical domains, a worst-case cost guarantee is preferred over the average cost bound~\cite{yang2021wcsac}. To achieve this, we use  CVaR~\cite{rockafellar2000optimization} instead of the expected value of cost to threshold the safety of a policy. 

We set cost limit $K = 10$ for all $\alpha$, that is, the expectation of the cost of the worst $\alpha * 100\%$ cases should be lower than $K$. We evaluate our method with different $\alpha$ shown with CoSHRL-0.9, CoSHRL-0.5, and CoSHRL-0.1. All experiments are averaged over 100 runs.

In Table~\ref{table:fourroom_stochastic_cost}, the results show that our method CoSHRL with $\alpha=0.9, 0.5$ satisfy the corresponding CVaR bound (columns C$\alpha$ shows the estimated average costs of the worst $\alpha * 100\%$ trajectories) while CoSHRL violates the CVaR bound ($K=10$) slightly with the tight level $\alpha=0.1$ because of the inherent approximation in distributional RL, namely that of discretization and truncation of long-tailed multinomial distribution. 
As $\alpha$ decreases, our method is more risk-averse so the percentage of trajectories that exceed the cost limit $K$ decreases (\% column), and cost and reward both improve. The statistical properties of the total cost incurred by CoSHRL under different risk level $\alpha$ are shown in Figure~\ref{fig:fourroom_stochastic_cost}.

\begin{table}[t]
{\small
    \centering
    \begin{tabular}{@{}lrrrrrr@{}}
    \toprule
             & EC            & C0.9          & C0.5          & C0.1             & ENR    & \% \\ \midrule
    CoSHRL-0.9 & 7.90          & \textbf{8.47} & 10.20         & 14.67            & 47.83 & 16\%      \\
    CoSHRL-0.5 & 6.68          & 7.31          & \textbf{9.20} & 13.22            & 48.58 & 11\%      \\
    CoSHRL-0.1 & 6.47          & 7.06          & 8.86          & \textbf{12.11}   & 48.60 & 7\%       \\
    SORB     & 8.06          & 8.97          & 11.54         & 15.14            & 52.98 & 30.5\%    \\ \bottomrule
    \end{tabular}
    }
    \caption{Different metrics of performance in the environment with stochastic cost: expected cost (EC), cost-CVaR-0.9 (C0.9), cost-CVaR-0.5 (C0.5), cost-CVaR-0.1 (C0.1), and expected negated reward (ENR)}
    \label{table:fourroom_stochastic_cost}
\end{table}

\noindent \textbf{Image-based Navigation with Obstacles and Hazards}:
Due to the lack of a constrained image-based environment, we design the Safe-ViZDoom environment in Figure~\ref{fig:ViZDoom} based on ViZDoom~\cite{Wydmuch2019ViZdoom}. The Safe-VizDoom environment is a labyrinth in the shape of a clover with a hazardous area in the middle, making it challenging due to the very narrow safe area in the middle. The agent can move North/South/East/West by a fixed distance, whereas states only consist of first-person visual perspective (3x160x120 dimension). The agent incurs a cost $c = 1$ for each step in the hazard, otherwise $c = 0$. The start is randomly placed in one of the four rooms, while the goal is randomly set in the opposite room.

We evaluate CoSHRL with different cost limits $K$ shown as CoSHRL-0, CoSHRL-4, and CoSHRL-8 in Figure~\ref{fig:ViZDoom_success_rate_reward} and Figure~\ref{fig:ViZDoom_cost} without retraining the low level RL agent.  Each result is the average over 100 random runs. We obtain similar results to other domains.  Figure~\ref{fig:ViZDoom_success_rate_reward} shows that CoSHRL achieves a high success rate ($>\!95$\%) in reaching the goal with varying cost limits. As the cost limit increases, CoSHRL obtains shorter paths (avg. negated reward), indicating that the agent ventures deeper into hazards. For avg. negated reward, CoSHRL outperforms the \emph{unconstrained} SORB and GRL for cost limit $K\!=\!8$, which is roughly the cost incurred by SORB and GRL in Figure~\ref{fig:ViZDoom_cost}. Figure~\ref{fig:ViZDoom_cost} shows the proportions of trajectories exceeding the cost limits of $K=0, 4, 8$ are 2\%, 4\%, 5\% respectively.  In comparison, unconstrained SORB and GRL achieve shorter path lengths (average negated reward) but incur cost over 8 in over half of their trajectories. The non-hierarchical GRL has a low success rate of 55\%, resulting in the agent getting stuck in corners.
\section{Discussion}
We introduced a \emph{constrained search} within the hierarchical RL approach. The RL agent is utilized to find paths between any two ``nearby'' states. Then, the constrained search utilizes the RL agent to reach far away goal states from starting states, while satisfying various types of constraints.  We were able to demonstrate the better scalability, theoretical soundness, and empirical utility of our approach, CoSHRL, over existing approaches for Constrained RL and Hierarchical RL. Next, we  discuss some limitations and future work.

Our work is based on the assumption that the low level RL agent has a high success rate in reaching each waypoint, even though there might be events such as action execution failure.  RL in general can handle action execution uncertainty (process noise) by observing the current (unexpected) state after an action failure and appropriately executing contingency actions from such observations. Thus, the low level RL will ultimately reach the local goal even though it might occasionally (with some probability) take more steps due to action execution failure. In extreme cases, due to poor generalization the low level RL can declare a state unreachable, even though the state might be reachable. This can sometimes result in no path being found to the final goal. However, this happens very rarely, which
is the main reason why the success rate of our method in the
test environments (Figures~\ref{fig:fourroom_success_rate_reward},~\ref{fig:ViZDoom_success_rate_reward}) are not exactly 100\%. A possible direction to improve this is for constrained RRT* to actively ask for retraining the low level agent; an active retraining paradigm could be an interesting future research direction.

\section{Acknowledgments}
This research/project is supported by the National Research Foundation Singapore and DSO National Laboratories under the AI Singapore Programme (AISG Award No: AISG2-RP-2020-017)
\bibliography{aaai24}

\begin{thebibliography}{40}
\providecommand{\natexlab}[1]{#1}

\bibitem[{Achiam et~al.(2017)Achiam, Held, Tamar, and Abbeel}]{achiam2017constrained}
Achiam, J.; Held, D.; Tamar, A.; and Abbeel, P. 2017.
\newblock Constrained policy optimization.
\newblock In \emph{International conference on machine learning}, 22--31. PMLR.

\bibitem[{Beker, Mohammadi, and Zamir(2022)}]{beker2022palmer}
Beker, O.; Mohammadi, M.; and Zamir, A. 2022.
\newblock PALMER: Perception-Action Loop with Memory for Long-Horizon Planning.
\newblock \emph{arXiv preprint arXiv:2212.04581}.

\bibitem[{Bellemare, Dabney, and Rowland(2023)}]{bdr2023}
Bellemare, M.~G.; Dabney, W.; and Rowland, M. 2023.
\newblock \emph{Distributional Reinforcement Learning}.
\newblock MIT Press.
\newblock \url{http://www.distributional-rl.org}.

\bibitem[{Chow et~al.(2017)Chow, Ghavamzadeh, Janson, and Pavone}]{chow2017risk}
Chow, Y.; Ghavamzadeh, M.; Janson, L.; and Pavone, M. 2017.
\newblock Risk-constrained reinforcement learning with percentile risk criteria.
\newblock \emph{The Journal of Machine Learning Research}, 18(1): 6070--6120.

\bibitem[{Chow et~al.(2018)Chow, Nachum, Duenez-Guzman, and Ghavamzadeh}]{chow2018lyapunov}
Chow, Y.; Nachum, O.; Duenez-Guzman, E.; and Ghavamzadeh, M. 2018.
\newblock A lyapunov-based approach to safe reinforcement learning.
\newblock \emph{Advances in neural information processing systems}, 31.

\bibitem[{Dietterich(2000)}]{dietterich2000hierarchical}
Dietterich, T.~G. 2000.
\newblock Hierarchical reinforcement learning with the MAXQ value function decomposition.
\newblock \emph{Journal of artificial intelligence research}, 13: 227--303.

\bibitem[{Eysenbach et~al.(2018)Eysenbach, Gupta, Ibarz, and Levine}]{eysenbach2018diversity}
Eysenbach, B.; Gupta, A.; Ibarz, J.; and Levine, S. 2018.
\newblock Diversity is all you need: Learning skills without a reward function.
\newblock \emph{arXiv preprint arXiv:1802.06070}.

\bibitem[{Eysenbach, Salakhutdinov, and Levine(2019)}]{eysenbach2019search}
Eysenbach, B.; Salakhutdinov, R.~R.; and Levine, S. 2019.
\newblock Search on the replay buffer: Bridging planning and reinforcement learning.
\newblock \emph{Advances in Neural Information Processing Systems}, 32.

\bibitem[{Fran{\c{c}}ois-Lavet et~al.(2018)Fran{\c{c}}ois-Lavet, Henderson, Islam, Bellemare, Pineau et~al.}]{franccois2018introduction}
Fran{\c{c}}ois-Lavet, V.; Henderson, P.; Islam, R.; Bellemare, M.~G.; Pineau, J.; et~al. 2018.
\newblock An introduction to deep reinforcement learning.
\newblock \emph{Foundations and Trends{\textregistered} in Machine Learning}, 11(3-4): 219--354.

\bibitem[{Gammell, Srinivasa, and Barfoot(2014)}]{gammell2014informed}
Gammell, J.~D.; Srinivasa, S.~S.; and Barfoot, T.~D. 2014.
\newblock Informed RRT*: Optimal sampling-based path planning focused via direct sampling of an admissible ellipsoidal heuristic.
\newblock In \emph{2014 IEEE/RSJ International Conference on Intelligent Robots and Systems}, 2997--3004. IEEE.

\bibitem[{Gattami, Bai, and Aggarwal(2021)}]{gattami2021reinforcement}
Gattami, A.; Bai, Q.; and Aggarwal, V. 2021.
\newblock Reinforcement learning for constrained markov decision processes.
\newblock In \emph{International Conference on Artificial Intelligence and Statistics}, 2656--2664. PMLR.

\bibitem[{Hernandez-Leal, Kartal, and Taylor(2019)}]{hernandez2019survey}
Hernandez-Leal, P.; Kartal, B.; and Taylor, M.~E. 2019.
\newblock A survey and critique of multiagent deep reinforcement learning.
\newblock \emph{Autonomous Agents and Multi-Agent Systems}, 33(6): 750--797.

\bibitem[{Jothimurugan et~al.(2021)Jothimurugan, Bansal, Bastani, and Alur}]{jothimurugan2021compositional}
Jothimurugan, K.; Bansal, S.; Bastani, O.; and Alur, R. 2021.
\newblock Compositional reinforcement learning from logical specifications.
\newblock \emph{Advances in Neural Information Processing Systems}, 34: 10026--10039.

\bibitem[{Kaelbling(1993)}]{kaelbling1993learning}
Kaelbling, L.~P. 1993.
\newblock Learning to achieve goals.
\newblock In \emph{IJCAI}, volume~2, 1094--8. Citeseer.

\bibitem[{Karaman and Frazzoli(2011)}]{karaman2011sampling}
Karaman, S.; and Frazzoli, E. 2011.
\newblock Sampling-based algorithms for optimal motion planning.
\newblock \emph{The international journal of robotics research}, 30(7): 846--894.

\bibitem[{Kim, Seo, and Shin(2021)}]{kim2021landmark}
Kim, J.; Seo, Y.; and Shin, J. 2021.
\newblock Landmark-guided subgoal generation in hierarchical reinforcement learning.
\newblock \emph{Advances in Neural Information Processing Systems}, 34: 28336--28349.

\bibitem[{Kim, Ahn, and Bengio(2019)}]{kim2019variational}
Kim, T.; Ahn, S.; and Bengio, Y. 2019.
\newblock Variational temporal abstraction.
\newblock \emph{Advances in Neural Information Processing Systems}, 32.

\bibitem[{Kulkarni et~al.(2016)Kulkarni, Narasimhan, Saeedi, and Tenenbaum}]{kulkarni2016hierarchical}
Kulkarni, T.~D.; Narasimhan, K.; Saeedi, A.; and Tenenbaum, J. 2016.
\newblock Hierarchical deep reinforcement learning: Integrating temporal abstraction and intrinsic motivation.
\newblock \emph{Advances in neural information processing systems}, 29.

\bibitem[{Levy et~al.(2017)Levy, Konidaris, Platt, and Saenko}]{levy2017learning}
Levy, A.; Konidaris, G.; Platt, R.; and Saenko, K. 2017.
\newblock Learning multi-level hierarchies with hindsight.
\newblock \emph{arXiv preprint arXiv:1712.00948}.

\bibitem[{Liang, Que, and Modiano(2018)}]{liang2018accelerated}
Liang, Q.; Que, F.; and Modiano, E. 2018.
\newblock Accelerated primal-dual policy optimization for safe reinforcement learning.
\newblock \emph{arXiv preprint arXiv:1802.06480}.

\bibitem[{Liu, Zhu, and Zhang(2022)}]{ijcai2022p770}
Liu, M.; Zhu, M.; and Zhang, W. 2022.
\newblock Goal-Conditioned Reinforcement Learning: Problems and Solutions.
\newblock In Raedt, L.~D., ed., \emph{Proceedings of the Thirty-First International Joint Conference on Artificial Intelligence, {IJCAI-22}}, 5502--5511. International Joint Conferences on Artificial Intelligence Organization.
\newblock Survey Track.

\bibitem[{Liu et~al.(2022)Liu, Cen, Isenbaev, Liu, Wu, Li, and Zhao}]{liu2022constrained}
Liu, Z.; Cen, Z.; Isenbaev, V.; Liu, W.; Wu, S.; Li, B.; and Zhao, D. 2022.
\newblock Constrained variational policy optimization for safe reinforcement learning.
\newblock In \emph{International Conference on Machine Learning}, 13644--13668. PMLR.

\bibitem[{Nachum et~al.(2018)Nachum, Gu, Lee, and Levine}]{nachum2018data}
Nachum, O.; Gu, S.~S.; Lee, H.; and Levine, S. 2018.
\newblock Data-efficient hierarchical reinforcement learning.
\newblock \emph{Advances in neural information processing systems}, 31.

\bibitem[{Neary et~al.(2022)Neary, Verginis, Cubuktepe, and Topcu}]{neary2022verifiable}
Neary, C.; Verginis, C.; Cubuktepe, M.; and Topcu, U. 2022.
\newblock Verifiable and compositional reinforcement learning systems.
\newblock In \emph{Proceedings of the International Conference on Automated Planning and Scheduling}, volume~32, 615--623.

\bibitem[{Pankayaraj and Varakantham(2023)}]{pankayaraj2023constrained}
Pankayaraj, P.; and Varakantham, P. 2023.
\newblock Constrained reinforcement learning in hard exploration problems.
\newblock In \emph{Proceedings of the AAAI Conference on Artificial Intelligence}, volume~37, 15055--15063.

\bibitem[{Ray, Achiam, and Amodei(2019)}]{ray2019benchmarking}
Ray, A.; Achiam, J.; and Amodei, D. 2019.
\newblock Benchmarking safe exploration in deep reinforcement learning.
\newblock \emph{arXiv preprint arXiv:1910.01708}, 7: 1.

\bibitem[{Rockafellar, Uryasev et~al.(2000)}]{rockafellar2000optimization}
Rockafellar, R.~T.; Uryasev, S.; et~al. 2000.
\newblock Optimization of conditional value-at-risk.
\newblock \emph{Journal of risk}, 2: 21--42.

\bibitem[{Ross(2014)}]{ross2014introduction}
Ross, S.~M. 2014.
\newblock \emph{Introduction to probability models}.
\newblock Academic press.

\bibitem[{Roza, Roscher, and G{\"u}nnemann(2023)}]{roza2023safe}
Roza, F.~S.; Roscher, K.; and G{\"u}nnemann, S. 2023.
\newblock Safe and Efficient Operation with Constrained Hierarchical Reinforcement Learning.
\newblock In \emph{Sixteenth European Workshop on Reinforcement Learning}.

\bibitem[{Satija, Amortila, and Pineau(2020)}]{satija2020constrained}
Satija, H.; Amortila, P.; and Pineau, J. 2020.
\newblock Constrained markov decision processes via backward value functions.
\newblock In \emph{International Conference on Machine Learning}, 8502--8511. PMLR.

\bibitem[{Sim{\~a}o, Jansen, and Spaan(2021)}]{simao2021alwayssafe}
Sim{\~a}o, T.~D.; Jansen, N.; and Spaan, M.~T. 2021.
\newblock AlwaysSafe: Reinforcement learning without safety constraint violations during training.
\newblock In \emph{Proceedings of the 20th International Conference on Autonomous Agents and MultiAgent Systems}. International Foundation for Autonomous Agents and Multiagent Systems.

\bibitem[{Sootla et~al.(2022)Sootla, Cowen-Rivers, Jafferjee, Wang, Mguni, Wang, and Ammar}]{sootla2022saute}
Sootla, A.; Cowen-Rivers, A.~I.; Jafferjee, T.; Wang, Z.; Mguni, D.~H.; Wang, J.; and Ammar, H. 2022.
\newblock Saut{\'e} rl: Almost surely safe reinforcement learning using state augmentation.
\newblock In \emph{International Conference on Machine Learning}, 20423--20443. PMLR.

\bibitem[{Stooke, Achiam, and Abbeel(2020)}]{stooke2020responsive}
Stooke, A.; Achiam, J.; and Abbeel, P. 2020.
\newblock Responsive safety in reinforcement learning by pid lagrangian methods.
\newblock In \emph{International Conference on Machine Learning}, 9133--9143. PMLR.

\bibitem[{Sutton, Precup, and Singh(1999)}]{sutton1999between}
Sutton, R.~S.; Precup, D.; and Singh, S. 1999.
\newblock Between MDPs and semi-MDPs: A framework for temporal abstraction in reinforcement learning.
\newblock \emph{Artificial intelligence}, 112(1-2): 181--211.

\bibitem[{Tessler, Mankowitz, and Mannor(2018)}]{tessler2018reward}
Tessler, C.; Mankowitz, D.~J.; and Mannor, S. 2018.
\newblock Reward constrained policy optimization.
\newblock \emph{arXiv preprint arXiv:1805.11074}.

\bibitem[{Wydmuch, Kempka, and Ja\'skowski(2019)}]{Wydmuch2019ViZdoom}
Wydmuch, M.; Kempka, M.; and Ja\'skowski, W. 2019.
\newblock {ViZDoom} {C}ompetitions: {P}laying {D}oom from {P}ixels.
\newblock \emph{IEEE Transactions on Games}, 11(3): 248--259.
\newblock The 2022 IEEE Transactions on Games Outstanding Paper Award.

\bibitem[{Yang et~al.(2021)Yang, Sim{\~a}o, Tindemans, and Spaan}]{yang2021wcsac}
Yang, Q.; Sim{\~a}o, T.~D.; Tindemans, S.~H.; and Spaan, M.~T. 2021.
\newblock WCSAC: Worst-Case Soft Actor Critic for Safety-Constrained Reinforcement Learning.
\newblock In \emph{AAAI}, 10639--10646.

\bibitem[{Yu, Xu, and Zhang(2022)}]{yu2022towards}
Yu, H.; Xu, W.; and Zhang, H. 2022.
\newblock Towards safe reinforcement learning with a safety editor policy.
\newblock \emph{Advances in Neural Information Processing Systems}, 35: 2608--2621.

\bibitem[{Zhang et~al.(2020)Zhang, Guo, Tan, Hu, and Chen}]{zhang2020generating}
Zhang, T.; Guo, S.; Tan, T.; Hu, X.; and Chen, F. 2020.
\newblock Generating adjacency-constrained subgoals in hierarchical reinforcement learning.
\newblock \emph{Advances in Neural Information Processing Systems}, 33: 21579--21590.

\bibitem[{Zhang, Vuong, and Ross(2020)}]{zhang2020first}
Zhang, Y.; Vuong, Q.; and Ross, K. 2020.
\newblock First order constrained optimization in policy space.
\newblock \emph{Advances in Neural Information Processing Systems}, 33: 15338--15349.

\end{thebibliography}

\end{document}